\documentclass[conference]{IEEEtran}
\IEEEoverridecommandlockouts
\usepackage{cite}
\usepackage{amsmath,amssymb,amsfonts}
\usepackage{algorithmic}
\usepackage{graphicx}
\usepackage{textcomp}
\usepackage{xcolor}
\def\BibTeX{{\rm B\kern-.05em{\sc i\kern-.025em b}\kern-.08em
    T\kern-.1667em\lower.7ex\hbox{E}\kern-.125emX}}
\begin{document}

\title{TRIP: Trainable Region-of-Interest Prediction for Hardware-Efficient Neuromorphic Processing on Event-based Vision
\thanks{This work was partially funded by research and innovation projects REBECCA (KDT JU under grant agreement No. 101097224), NeuroKIT2E (KDT JU under grant agreement No. 101112268), and NimbleAI (Horizon EU under grant agreement 101070679).\\
$^{\dag}$ Corresponding author. Email: guangzhi.tang@maastrichtuniversity.nl\\
$^*$ Code: https://github.com/ERNIS-LAB/TRIP}
}

\author{\IEEEauthorblockN{Cina Arjmand$^{1}$, Yingfu Xu$^{1}$, Kevin Shidqi$^{1}$, Alexandra F. Dobrita$^{1}$, Kanishkan Vadivel$^{1}$,\\Paul Detterer$^{1}$, Manolis Sifalakis$^{1}$, Amirreza Yousefzadeh$^{2}$ and Guangzhi Tang$^{3,\dag}$}

\textit{$^{1}$ imec, Eindhoven, The Netherlands $^{2}$ EEMCS, University of Twente, Enschede, The Netherlands}\\

\textit{$^{3}$ DACS, Maastricht University, Maastricht, The Netherlands}

}

\maketitle

\begin{abstract}
Neuromorphic processors are well-suited for efficiently handling sparse events from event-based cameras. However, they face significant challenges in the growth of computing demand and hardware costs as the input resolution increases. This paper proposes the Trainable Region-of-Interest Prediction (TRIP)$^*$, the first hardware-efficient hard attention framework for event-based vision processing on a neuromorphic processor. Our TRIP framework actively produces low-resolution Region-of-Interest (ROIs) for efficient and accurate classification. The framework exploits sparse events' inherent low information density to reduce the overhead of ROI prediction. We introduced extensive hardware-aware optimizations for TRIP and implemented the hardware-optimized algorithm on the SENECA neuromorphic processor. We utilized multiple event-based classification datasets for evaluation. Our approach achieves state-of-the-art accuracies in all datasets and produces reasonable ROIs with varying locations and sizes. On the DvsGesture dataset, our solution requires $46\times$ less computation than the state-of-the-art while achieving higher accuracy. Furthermore, TRIP enables more than $2\times$ latency and energy improvements on the SENECA neuromorphic processor compared to the conventional solution.
\end{abstract}

\section{Introduction}
\label{sec:intro}

Low-power and low-latency event-based vision is uniquely suited for edge applications. Given the efficiency of sensing, developing equally efficient processing becomes crucial for optimizing the performance of edge solutions. Since the event-based camera inherently generates sparse data, exploiting this sparsity is essential for enhancing the processing efficiency. Neuromorphic computing offers event-driven solutions to process sparse data streams efficiently, making it a natural fit for event-based vision \cite{yik2023neurobench,xu18optimizing}. However, with the growing resolution of event-based cameras, neuromorphic computing faces computing and hardware cost challenges \cite{yik2023neurobench}. These challenges are further amplified when employing Convolutional Neural Networks (CNNs), as the computational expenses and on-chip memory demands for processing CNNs on neuromorphic processors increase with input resolution \cite{rueckauer2022nxtf}.

\begin{figure}[t]
  \centering
   \includegraphics[width=1\linewidth]{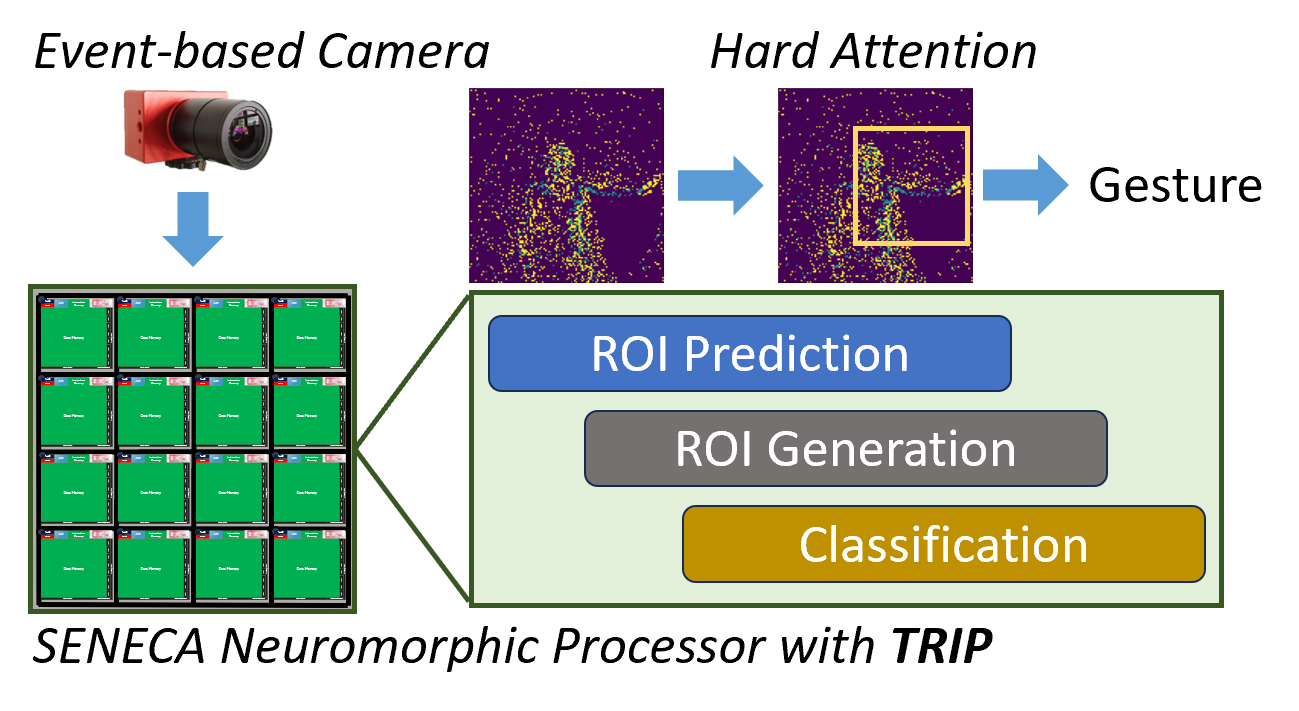}

   \caption{Overview of TRIP performing event-based vision classification on the SENECA neuromorphic processor.}
   \label{fig:onecol}
\end{figure}

To address the challenges of high-resolution visual processing, one approach is the hard attention algorithm, which selectively focuses on regions of an input image for processing \cite{Larochelle,mnih2014recurrent}. Compared to uniformly downsampling the entire input, the hard attention mechanism actively chooses regions of interest (ROI) with more critical information, improving accuracy while limiting the computing and memory costs of network processing. However, hard attention algorithms require an additional neural network to predict the ROI accurately. This demands sophisticated training methods and introduces additional overheads for visual processing. Therefore, the benefits gained from processing a reduced-dimension ROI can be offset by the high costs of ROI prediction \cite{chai2019patchwork}. The trade-off becomes particularly pronounced as the complexity of the scene increases, potentially negating the efficiency gains in ROI processing \cite{elsayed2019saccader}.

Interestingly, the inherent sparsity of event-based vision reduces the information density of scenes \cite{cohen2018spatial}, which can potentially mitigate the hard attention overhead on ROI prediction. This characteristic opens up opportunities for efficient event-based vision processing, especially when hard attention is integrated with neuromorphic processors. By reducing input dimensionality, the hard attention algorithm can significantly reduce the computational and memory demands of CNNs on neuromorphic processors. Moreover, the event-driven processing further diminishes the latency and energy overhead associated with hard attention when utilizing CNNs with sparse activation \cite{tang2023open}. This synergy opens prospects for tailoring hard attention algorithms on the neuromorphic processor.

In this paper, we propose the Trainable Region-of-Interest Prediction (TRIP) framework for hardware-efficient event-based vision processing on the neuromorphic processor. Our TRIP framework performs efficient ROI prediction with low-resolution event streams and supports end-to-end training by employing differentiable truncated Gaussian kernels (tGK) for ROI generation. We introduced hardware-aware optimizations for TRIP to improve the algorithm's hardware efficiency without sacrificing accuracy. We implemented the hardware-optimized TRIP algorithm on the SENECA neuromorphic processor \cite{tang2023seneca} and evaluated our method on event-based classification datasets \cite{amir2017low,muller2023aircraft}. Our method achieves state-of-the-art accuracies while reducing the computation cost by 46$\times$ compared to the state-of-the-art efficient algorithm \cite{subramoney2023efficient}. Compared to neuromorphic solutions on Intel's Loihi and IBM's TrueNorth neuromorphic processors \cite{rueckauer2022nxtf,amir2017low,massa2020efficient}, our TRIP-based solution significantly reduces the area and energy consumption while having higher accuracy.

%------------------------------------------------------------------------
\section{Related Works}

\subsection{Hard Attention Visual Processing}

Hard attention strategies for restricting computations by directing image processing towards relevant regions of input space have long been explored in computer vision. Early models analyze low-level image features to predict regions of high saliency based on variations in pixel intensity \cite{Itti}. Later works increasingly emphasized the task of salient region prediction as an action selection policy \cite{mnih2014recurrent}, iteratively improving predictions over time. Reinforcement learning (RL) algorithms have been adopted in hard attention to learn the optimal policy for placing a sensor with limited bandwidth on a given input region \cite{elsayed2019saccader,chai2019patchwork,kong2022efficient}. While RL-based hard attention is effective, the training complexity poses major challenges for hardware-aware optimization.

Deep Recurrent Attention Writer (DRAW) uses recurrent units within a variational autoencoder to iteratively predict salient regions of input images \cite{gregor2015draw}. Importantly, DRAW uses a differentiable mechanism for generating an ROI with Gaussian kernels, enabling end-to-end backpropagation training without using RL. Neuromorphic DRAW applied the differentiable cropping of DRAW to event-based classification tasks to improve accuracy by filtering out irrelevant events \cite{cannici2019attention}. Our TRIP framework leverages DRAW's Gaussian kernels to facilitate differentiable hard attention while introducing hardware-efficient algorithm designs for event-based vision processing on neuromorphic processors.

\subsection{Event-based CNN and SENECA Neuromorphic Processor}

Event-based CNN, trained by specialized activation regularization methods, has high activation sparsity within each network layer \cite{zhu2023star}. SENECA is a multi-core embedded digital neuromorphic processor specialized in processing event-based CNNs \cite{tang2023seneca}. It performs event-driven computation that exploits the sparsity in sensory inputs and network activations. Additionally, it executes data-flow processing across cores, increasing the parallelism of network processing and diminishing the memory cost for neural activations. Event-driven depth-first convolution is a unique scheduling method SENECA supports for event-based CNNs \cite{xu18optimizing}. The method prioritizes the network's layer dimension by consuming neural activation events right after their generation. Therefore, it maximizes the neuromorphic processor's benefits on parallelism and latency. Our TRIP framework with event-based CNN maximizes the hardware efficiency of hard attention on SENECA by exploiting the hardware advantages.

%------------------------------------------------------------------------
\section{Method}

\subsection{TRIP: Trainable Region-of-Interest Prediction}

\begin{figure*}
    \centering
    \includegraphics[width=0.86\linewidth]{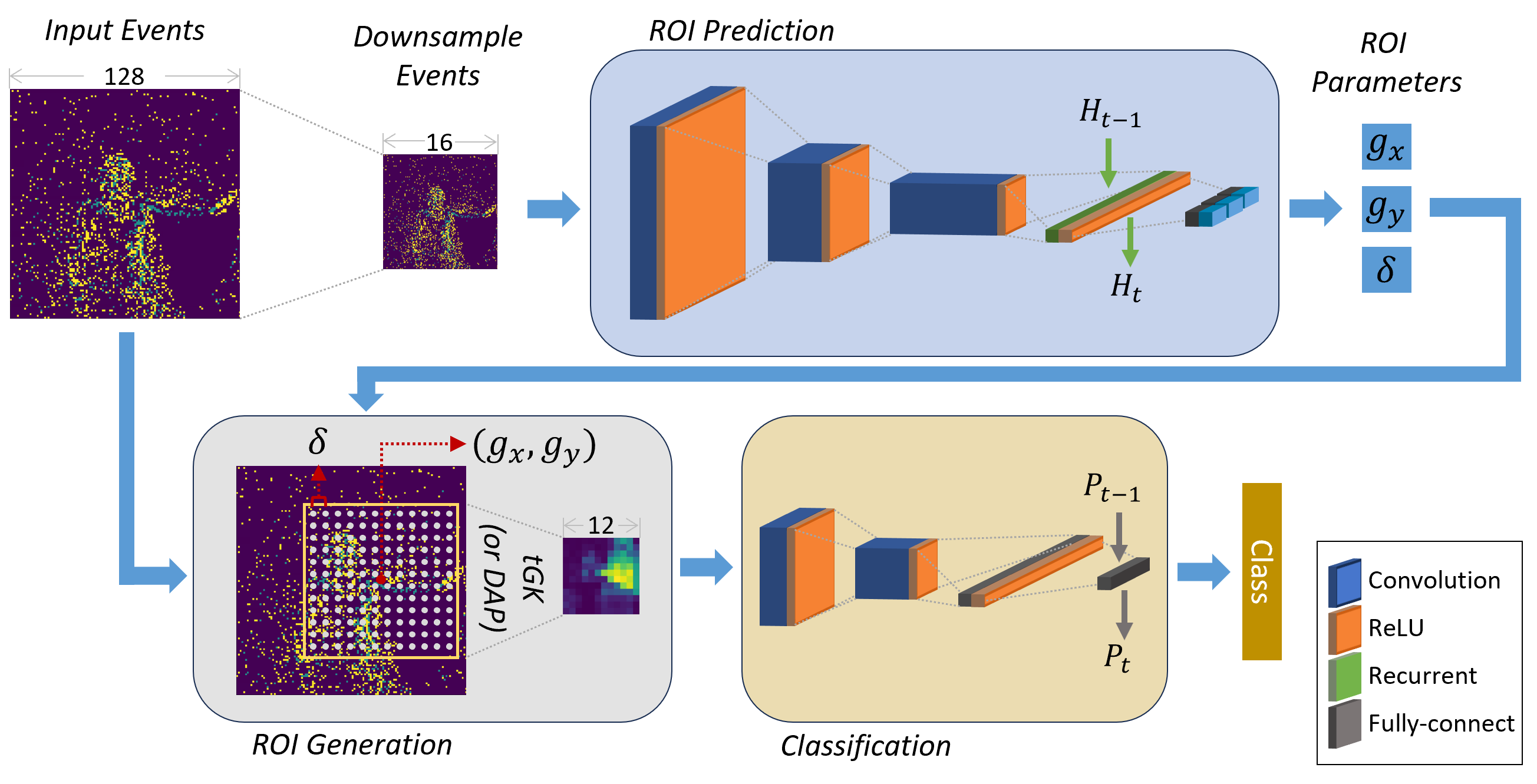}
    \caption{Processing pipeline of TRIP for the event-based gesture recognition task. The downsampled events are fed into the ROI prediction event-based CNN to predict the ROI parameters. The ROI generation module uses the parameters to create the ROI fed into the classification event-based CNN. $H_t$ is the output of the ReLU recurrent unit, and $P_t$ is the output for processing events in timebin $t$.}
    \label{fig:method}
\end{figure*}

We propose the Trainable Region-of-Interest Prediction (TRIP) framework for efficient event-based classification. It uses hard attention within an event-driven neuromorphic processing pipeline. The framework efficiently classifies event streams using an actively generated ROI that is predicted from the input events. An ROI's receptive field covers a small region of the event-based camera's field of view. As shown in Figure \ref{fig:method}, our TRIP framework consists of three subsequent components: ROI prediction, ROI generation, and classification. The ROI prediction component consists of an event-based CNN that determines the location and receptive field of the ROI. It predicts the ROI parameters using a downsampled low-resolution input, reducing the processing overhead of ROI prediction. The ROI generation component generates the cropped ROI using the predicted parameters. It uses an $N\times N$ grid of differentiable truncated Gaussian kernels (tGK) to produce a fixed $N\times N$ output from a varying-size receptive field. This ensures consistently low processing cost of classification. Moreover, we introduce dynamic average pooling (DAP) to replace tGK for efficient inference on the embedded neuromorphic processor. The classification component consists of an event-based CNN that performs classification on the ROI. The entire framework is differentiable, allowing it to be trained end-to-end. For efficient computing on SENECA, we increase the activation sparsity of the event-based networks during training.

\subsection{ROI Prediction}

The ROI prediction component produces the ROI parameters based on the downsampled input events from max-pooling. The ROI prediction network outputs three scalar values. These values are decoded to determine the ROI location and receptive field as follows,
\begin{equation}
    g_x = \frac{A}{2}\cdot (tanh(\hat{g_x})+1)
\end{equation}
\begin{equation}
    g_y = \frac{B}{2}\cdot  (tanh(\hat{g_y})+1)
\end{equation}
\begin{equation}
    \delta = S \cdot (sigmoid(\hat{\delta})+1)
\end{equation}
where $\hat{g_x},\hat{g_y},\hat{\delta}$ are the raw scalar outputs from the ROI prediction network, ($g_x,g_y$) is the center location of the overall receptive field formed by the $N\times N$ grid of tGK, $\delta$ is the distance between two adjacent tGK, $A$ and $B$ are the image width and height, and $S$ is a distance scaling factor. The $g_x$ and $g_y$ parameters always initialize in the center of the image at the start of training and can move across the entire input space. The variable $\delta$ allows control over the size of the receptive field of the ROI.

\subsection{ROI Generation}

We employed tGK for ROI generation, an efficient variation of the method introduced in DRAW \cite{gregor2015draw}. The ROI generation component generates the $N\times N \times 2$ fixed-resolution input ROI event streams for classification, in which $N$ is the width and height of the ROI and $2$ is the polarity channel. The component uses $N\times N$ differentiable tGK to compute the ROI during training. The 2D mean positions of the tGK are computed according to the predicted center location of the overall receptive field as follows, 
\begin{equation}
   \mu_{x}^{i}= g_x + (i - \frac{N}{2} - 0.5) \cdot \delta,\ i \in [0, N-1]
   \label{eq:grid_points}
\end{equation}
where $\mu_{x}^{i}$ is the mean x-axis position of the tGK on the $i^{th}$ column. The mean y-axis position of the tGK can be computed using the same equation with $g_y$. Eventually, each tGK has a two-dimensional mean position $(\mu_{x}^{i},\mu_{y}^{j})$, in which $i$ and $j$ are the column and row index. Here, we assume $N$ is an even number.

By knowing the mean positions of the tGK, we compute the weight of the tGK corresponding to each pixel location of the input events as follows,
\begin{equation}
F_{x}^i[n] = 
\begin{cases}
  \exp({\frac{(n - \mu_{x}^{i})^2}{2\sigma}}) & \text{for } n \in [\mu_{x}^{i} - \frac{\theta}{2},\  \mu_{x}^{i} + \frac{\theta}{2}]\\    
  0  & \text{otherwise}
\end{cases}
\label{eq:fx}
\end{equation}
where $F_{x}^i[n]$ is the x-dimension weight component of tGK on the $i^{th}$ column corresponding to pixel locations at column $n$, $\sigma$ is the variance which is a pre-defined parameter, and $\theta$ is the size of the tGK with non-zero weights. The y-dimension weight component is computed in a similar manner.

Each ROI input event value to the classification network is computed by the corresponding tGK as follows,
\begin{equation}
 v_{(x_i, y_j)} =  \mathbf{F_{x}^i}\cdot \mathbf{I}\cdot \mathbf{{F_{y}^j}}
\label{eq:vxixj}
\end{equation}
where $v_{(x_i, y_j)}$ is the value of the event at location $(x_i, y_j)$ of the $N\times N \times 2$ input to the classification network, $\mathbf{F_{x}^i}$ and $\mathbf{{F_{y}^j}}$ are the weights for the corresponding tGK, and $\mathbf{I}$ are the binned raw input events from one polarity. The two polarity channels of the classification inputs are computed using the same equation.

Compared to Gaussian kernels introduced by DRAW, our tGK significantly reduces the computation required for ROI generation while maintaining differentiable. Specifically, our adoption of tGK reduces the computational complexity of ROI generation from $O(AB)$ to $O(\theta^2)$ by skipping the pixel locations with insignificant weights. Since $\theta$ is at least ten times smaller than $A$ and $B$ in practice, the tGK can be orders of magnitude more efficient than Gaussian kernels.

\subsection{Hardware-Efficient Dynamic Average Pooling}

The tGK can be accelerated by customized application-specific integrated circuit (ASIC) designs. However, its efficiency is hard to achieve on the embedded CPU within our targeted neuromorphic processor. Though the number of computations is small, the overheads of locating the non-zero elements and performing weighted operations for tGK are substantial. Firstly, assigning each input event to multiple overlapping kernels requires a complex implementation to avoid iterating all kernels, introducing significant overhead to the instruction memory. Secondly, the distance between the kernel center and the event location must be computed for each assigned event, bringing additional overhead on computation. Hence, tGK on an embedded core with limited instruction memory and compute capability is not feasible.  

To mitigate the problem, we introduce Dynamic Average Pooling (DAP) as a hardware-efficient alternative for ROI generation during inference on the embedded neuromorphic processor. The DAP replaces the Gaussian kernels with simple non-overlapping average poolings. The kernel size of the average pooling changes dynamically based on the size of the overall receptive field the ROI corresponds to. We compute the range of the overall receptive field using the ROI parameters from the ROI prediction component as follows,
\begin{equation}
    x_{max} = g_x + (\frac{N}{2} - 0.5) \cdot \delta + \frac{\theta}{2}
\label{eq:bounds1}
\end{equation}
\begin{equation}
    x_{min} = g_x - (\frac{N}{2} + 0.5) \cdot \delta - \frac{\theta}{2}
\label{eq:bounds2}
\end{equation}
and the dynamic kernel size of each average pooling in the DAP is computed as follows,
\begin{equation}
    k_{DAP} = (x_{max} - x_{min}) / N
\end{equation}
where $x_{max}$ and $x_{min}$ define the range of the receptive field on the x-axis of the raw input space. The range on the y-axis can be computed using the same equations with $g_y$ and the receptive field is square.

The embedded implementation of DAP is simple. A closed-form equation exists to compute the sole corresponding kernel of each input event, and the ROI generation is distance invariant. However, the ROI parameters in DAP are not differentiable. Therefore, we first train the networks with tGK, and then fine-tune the classification network with a fixed ROI prediction network and DAP.

\subsection{Hardware-Aware Event-based CNN}
\label{method:hadwaretraining}

The event-driven neuromorphic processor exploits activation sparsities in neural networks by only processing non-zero activations. Therefore, input to each layer for synaptic operation is supposed to be as sparse as possible. To maximize the efficiency of TRIP on the event-driven neuromorphic processor, we adopt event-based CNNs for ROI prediction and classification. Unlike regular CNN, our event-based convolutional layer for the neuromorphic processor performs BatchNorm and MaxPool before the ReLU activation, outputting sparse events straight to the subsequent layer for synaptic integration. Moreover, we used ReLU function in the vanilla RNN for sparse recurrent processing. Furthermore, we perform hardware-aware optimizations on the event-based CNNs. The optimizations comprise sparsity-aware and quantization-aware training, reducing projected computation cost and memory requirement on the hardware.

To increase the activation sparsity of event-based CNNs in TRIP, we adopt the $L1$ regularization loss \cite{zhu2023star} on the activation values of the layers that have ReLU as the activation function. The loss encourages the network to reduce the activation values so as to have fewer non-zero activations and increase the sparsity. Additionally, we use quantization-aware training to reduce weight precision to 4 bits \cite{xu18optimizing}. There is a shared power-of-two scaling factor $s$ for all the weights of the same layer. During on-chip computation, a weight value is obtained by multiplying the saved 4-bit integer with $2^s$. The quantized parameters reduce the on-chip memory required for network deployments and the computation cost of synaptic integration.

%end method section

\begin{table*}
\centering
\caption{Performance comparisons on the DvsGesture dataset.}
\begin{tabular}{llllll}
\hline
Architecture & Input Resolution & Param & Effective MACs  & Accuracy [\%] & Accuracy [\%]\\ 
 & & & (Single Timebin) & (mean $\pm$ \text{ std}) & (Maximum)\\
\hline
LSTM \cite{he2020comparing} & 32$\times$32  & 7.4M         & 3.9M        & --             & 86.8  \\
AlexNet+LSTM\cite{innocenti2021temporal} & 128$\times$128     & 8.3M & 601.3M                 & --             & 97.7                    \\ 
CNN+EGRU \cite{subramoney2023efficient} & 128$\times$128     & 4.8M     & 80.6M             & 97.3$\pm$ 0.4            & 97.8                    \\ 
ConvLIAF \cite{wu2022liaf} & 32$\times$32 & \textbf{0.22M} & 113.3M & -- & 97.6 \\ \hline
TRIP (Ours) & 16$\times$16+12$\times$12      & 0.46M   & \textbf{1.75M}          & \textbf{97.6} $\pm$ \textbf{0.5}               & \textbf{98.6}                   \\ \hline
\end{tabular}
 \label{tb:DVS}
\end{table*}

\begin{figure*}[h]
  \centering
   \includegraphics[width=0.90\linewidth]{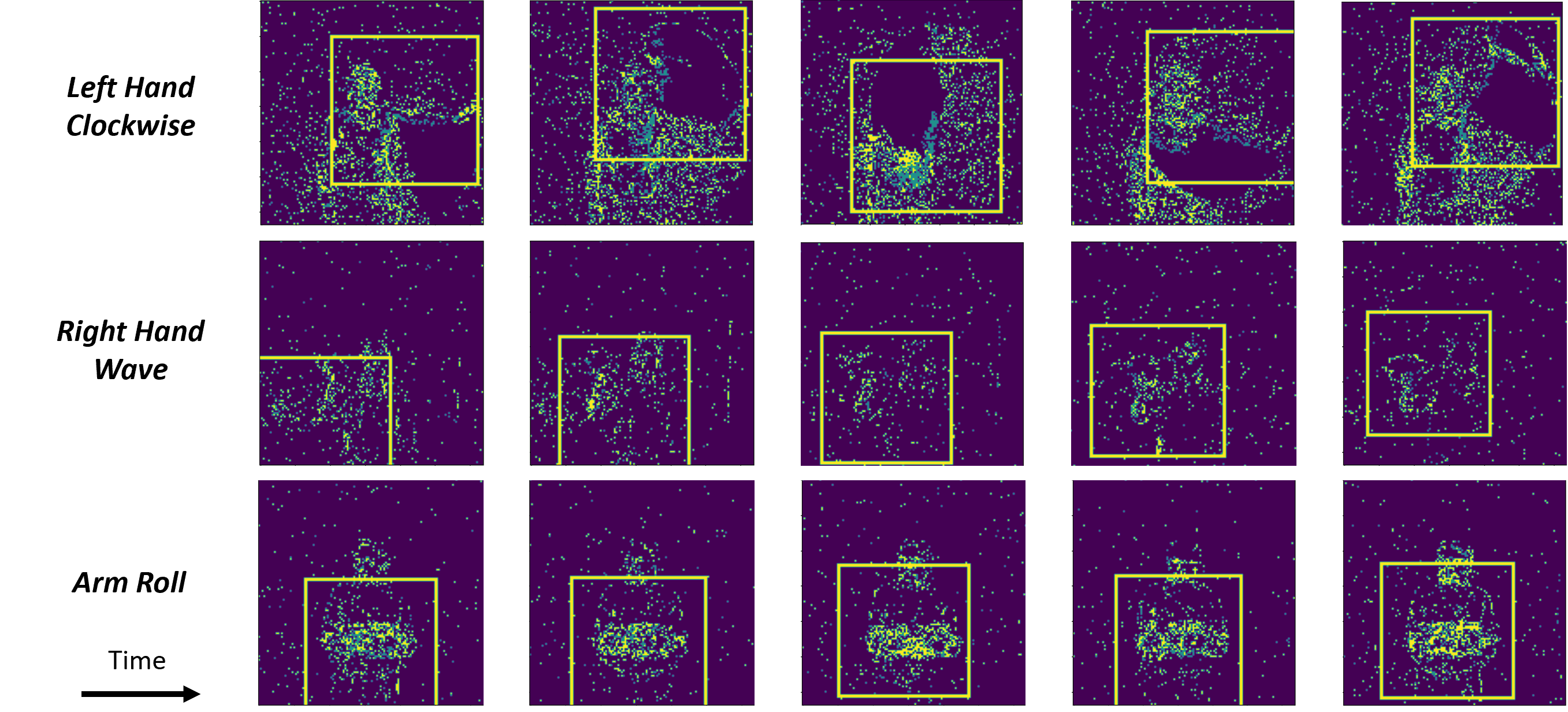}

   \caption{Visualization of ROI's receptive fields for different gestures in the DvsGesture dataset. The receptive fields include the pixels involved in the ROI generation. They have superimposed on top of the timebinned event streams as a yellow rectangle.}
   \label{fig:DVS}
\end{figure*}

\section{Experiment and Results}

We benchmarked the performance of TRIP on challenging event-based classification datasets and with the SENECA neuromorphic processor. Firstly, we experimented on the widely used DvsGesture dataset \cite{amir2017low} to demonstrate the effectiveness of TRIP in terms of accuracy, model size, and algorithmic computing cost. Secondly, we used the Marshalling Signals gesture recognition dataset \cite{muller2023aircraft} to evaluate TRIP's robustness towards samples at varying distances from the event-based camera. Our results demonstrate that the ROI prediction is dynamically adaptable to varying distances. Thirdly, we synthetically generated a noisy, high-resolution event-based dataset based on N-MNIST \cite{orchard2015converting} with digits in varying sizes and locations. Using this dataset, we validated the effectiveness and overhead of TRIP compared to baselines with similar cost or accuracy. Finally, we implemented our hardware-optimized TRIP algorithm on the SENECA neuromorphic processor and measured the energy, latency, and effective area of the solution.

\subsection{DvsGesture Dataset}

Gesture recognition is an ideal task for evaluating approaches with hard attention, as a compact region of the input can provide sufficient information for classification. The DvsGesture dataset \cite{amir2017low} enables us to compare our method with other state-of-the-art solutions on the task of gesture recognition with an event-based camera.

\subsubsection{Dataset and Network Overview}

The dataset is recorded using the DVS128 event-based camera with $128\times128$ resolution. It consists of 11 gesture classes in 1176 training sessions and 288 testing sessions. Each session includes a subject repeatedly performing the same gesture. We preprocessed each session using SpikingJelly \cite{fang2023spikingjelly} into an event sample of 32 timebins, in which the events from the same pixel location are accumulated together within each timebin. We performed data augmentation during training to randomly scale, rotate, and spatially shift training samples. We downsampled the input resolution to $16\times16$ for ROI prediction. The ROI prediction network comprises three convolutional layers, a ReLU recurrent layer, and an output layer. We used $12\times12$ tGKs to generate the ROI input for classification. The classification network comprises two convolutional layers, a fully-connected hidden layer, and an output layer.

\subsubsection{Results}

We compared accuracies, number of parameters, and effective MAC operations with other state-of-the-art methods in Table \ref{tb:DVS}. The effective MAC counts the averaged non-zero multiply-accumulate operations within all components of TRIP for processing a single timebin of the event stream, reflecting the computing cost on event-driven neuromorphic processors. Our TRIP framework achieves state-of-the-art accuracy while reducing the effective MAC by $46\times$ compared to the lowest among the other state-of-the-art approaches. TRIP achieves tremendous computational efficiency gains through two key differentiators: firstly, by operating on considerably lower input resolutions compared to other CNN-based methods, and secondly, by utilizing less complex network architectures while processing a reduced input space with less irrelevant information.

We visualized the receptive fields used for generating ROIs for classification in Figure \ref{fig:DVS}. The visualization helps for interpreting the decision process of TRIP and further explains the reason behind TRIP's efficiency advantage. By visually inspecting the samples, we can see the ROI prediction network learns to track the gestures intelligently and focus on salient regions of the input space. For example, in the ``left hand clockwise" gesture, the ROI's receptive field tracks the arm's movement, making the classification network easier to make a decision.

\subsection{Marshalling Signals Dataset}

The Marshalling Signals dataset \cite{muller2023aircraft} is more recent, less explored, and more difficult than DvsGesture. The dataset presents gestures at multiple distances from the event-based camera. Therefore, it allows us to further test the ROI prediction, particularly its ability to adjust to varying sizes.

\subsubsection{Dataset and Network Overview}

The Marshalling Signals dataset \cite{muller2023aircraft} is recorded using the DAVIS 346 event-based camera with $346\times224$ resolution. It contains 10 gesture classes in 11,040 training samples and 930 testing samples. Each sample is one gesture presented in a 960 ms timebin. Each gesture is presented in 8 evenly spaced distances from the camera ranging from 1.5m to 4.5m. We adopt the same network architectures as the DvsGesture task with a higher dimension ReLU recurrent layer in the ROI prediction network. We downsampled the input solution to $43\times28$ for ROI prediction and used $12\times12$ tGKs for ROI generation.

\subsubsection{Results}

\begin{figure}[t]
  \centering
   \includegraphics[width=1\linewidth]{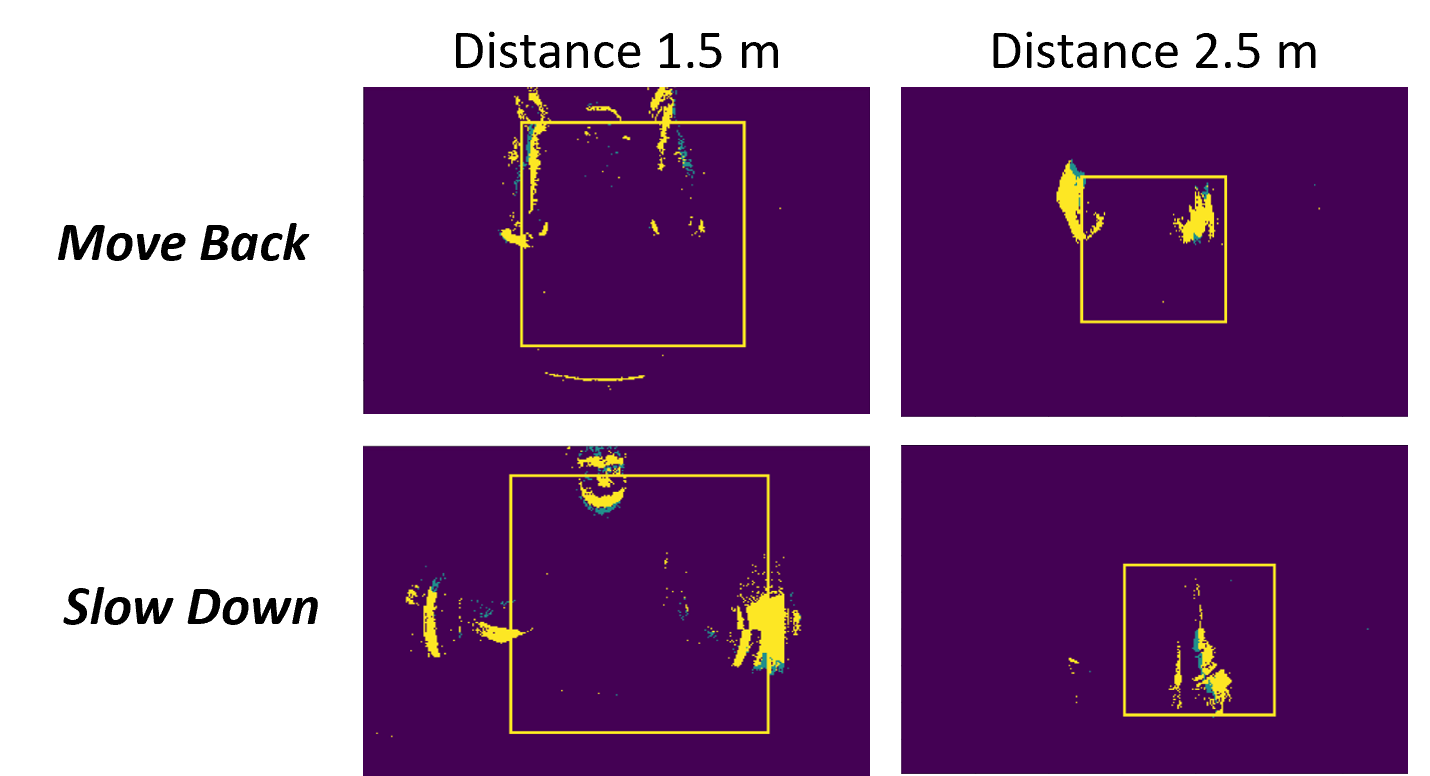}
   \caption{Visualization of ROI's receptive fields for gestures performed at different distances in the Marshalling Signals dataset.}
   \label{fig:MS}
\end{figure}

We compared the performance of our model with the previous results in Table \ref{tb:MS}. Since \cite{muller2023aircraft} uses regular CNN architectures, we used FLOPs as an efficiency metric, without considering the activation sparsities in our event-based CNNs. Our TRIP framework achieves better accuracy while reducing the FLOPs by $18\times$ compared to EfficientNet \cite{tan2019efficientnet}. By visualizing ROI's receptive fields for different distances in Figure \ref{fig:MS}, we show that the ROI prediction can adjust the ROI size for classification to include only the relevant region of the input space.

\begin{table}
\centering
\caption{Performance comparisons on the Marshalling Signals dataset.}
\begin{tabular}{lllll}
\hline
Architecture & Param & FLOPs &  Accuracy [\%] \\ \hline
ResNet18 \cite{muller2023aircraft}      & 11.7M      &     1810M      & 74.6                           \\ 
EfficientNet-B1 \cite{muller2023aircraft}      & 7.794M  &     690M           & 82.6                             \\  
\hline
TRIP (Ours)       & \textbf{4.13M}  &     \textbf{37.0M}      & \textbf{83.6}                               \\ \hline  
\end{tabular}
 \label{tb:MS}
\end{table}

\subsection{Synthetic Dataset based on N-MNIST}

To study the effects of the reduced input resolutions and the hard attention overheads in a controlled setup, we synthetically generated a dataset based on the N-MNIST dataset \cite{orchard2015converting}. The generated dataset enables us to test the performance of TRIP under different input resolutions and structured event noises.

\subsubsection{Dataset Generation and Network Overview}

We generated the synthetic N-MNIST dataset by randomly scaled event streams of $34\times34$ resolution N-MNIST digits on arbitrary locations of a $128\times128$ canvas. The scaling factor for each sample is randomly selected between 1 to 2. We add structured event noises by randomly selecting 8 other digits from the dataset, cropping a random $8\times8$ subsection of each digit, and placing the subsections in random locations on the canvas. Figure \ref{fig:nmnist} shows some examples of the generated samples. The synthetic dataset has the same number of samples as the original N-MNIST dataset, including 60,000 training and 10,000 testing samples.

We used the same network architectures in TRIP as the DvsGesture task but with only 2 convolutional layers for the ROI prediction network. We used $12\times12$ tGKs for ROI generation. The baseline networks have the same number of layers as TRIP, which comprises 4 convolutional layers, a ReLU recurrent layer, a fully-connected layer, and an output layer. We tested different input resolutions for the baselines and TRIP's ROI prediction, including $16\times16$, $32\times32$, and $64\times64$. The baseline networks have varying layer dimensions based on the input resolution.

\subsubsection{Results}

\begin{figure}[t]
  \centering
   \includegraphics[width=1\linewidth]{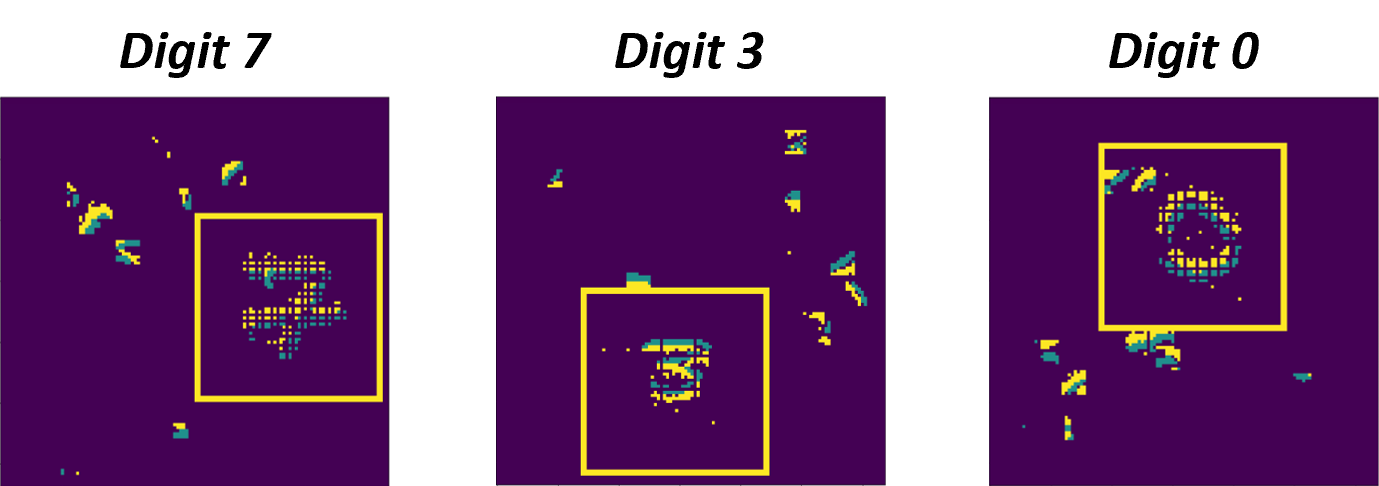}

   \caption{Example of synthetic N-MNIST samples (from left to right: digit 7, 3, and 0), showing ROI generated by network.}
   \label{fig:nmnist}
\end{figure}

We compared the performance of our model with the baseline models on different input resolutions in Table \ref{tb:NMNIST}. Comparing the baseline networks using one level higher input resolutions ($16\times16\rightarrow 32\times32$ and $32\times32\rightarrow 64\times64$), TRIP achieves higher or similar accuracies with reductions in FLOPs. This shows the low input resolution required by TRIP to maintain high accuracy compensates for the hard attention overheads introduced by the ROI prediction and generation. Moreover, the visualization results in Figure \ref{fig:nmnist} shows the ROI prediction network can handle inputs with structured noises which share similar features with the digits and hard to differentiate in low resolution.

\begin{table}
\centering
\caption{Performance comparisons on synthetic N-MNIST dataset.}
\begin{tabular}{llll}

\hline
Architecture & Param & FLOPs & Accuracy [\%] \\
 & & & (mean $\pm$ \text{ std})\\
\hline
Baseline (16x16)    & 0.31M   &   6.0M      & 71.8$\pm$ 2.3              \\
Baseline (32x32)    & 0.67M    &   24.4M     & 93.0$\pm$ 0.6                               \\ 
Baseline (64x64)      & 0.67M  &   57.4M     & 96.2$\pm$ 0.9                            \\ 
\hline
TRIP (16x16)       & 0.30M     & 16.0M       & 95.4 $\pm$ 0.4                  \\
TRIP (32x32)       & 0.65M    &  28.0M       & 96.1 $\pm$ 0.3                        \\ 
\hline
\end{tabular}
 \label{tb:NMNIST}
\end{table}

\subsection{Neuromorphic Processor Deployment}

\begin{table*}
\centering
\caption{Comparison with state-of-the-art neuromorphic implementations on the DvsGesture dataset.}
\begin{tabular}{lcccccccccc}
\hline
 & & & & \multicolumn{1}{c|}{} & \multicolumn{3}{c|}{Single Timebin} & \multicolumn{3}{c}{Multiple Timebins} \\ 
\hline
Hardware & Solutions & Technology & Core & \multicolumn{1}{c|}{Area} & Latency  &  $E_{inf}$ & \multicolumn{1}{c|}{Accuracy} & Latency  &  $E_{inf}$ & Accuracy\\ 
         &      & &   [\#]  & \multicolumn{1}{c|}{[mm$^2$]} & [ms] & [uJ] & \multicolumn{1}{c|}{[\%]} & [ms] & [uJ] & [\%]\\
\hline
Loihi \cite{davies2018loihi}      & Spiking CNN \cite{rueckauer2022nxtf} & Intel 14 nm & $>$20 & \multicolumn{1}{c|}{$>$8.20} & 11  & -- & \multicolumn{1}{c|}{89.6} & -- & -- & --\\
Loihi \cite{davies2018loihi}     & Spiking CNN \cite{massa2020efficient} & Intel 14 nm & 59 & \multicolumn{1}{c|}{24.19} & --  & -- & \multicolumn{1}{c|}{--} & \textbf{22.0} & 2731 & 96.2\\
TrueNorth \cite{akopyan2015truenorth}  & Spiking CNN \cite{amir2017low} & Samsung 28 nm & 3838 & \multicolumn{1}{c|}{383.8} & --  & -- & \multicolumn{1}{c|}{\textbf{91.8}} & 104.6 & 18702 & 94.6\\   
SENECA \cite{tang2023seneca} & Event-based CNN & GF FDX 22 nm & 7 & \multicolumn{1}{c|}{\textbf{3.29}} & --  & -- & \multicolumn{1}{c|}{--} & 78.9 & 1069.2 & 97.3\\
\hline
SENECA \cite{tang2023seneca} & TRIP & GF FDX 22 nm & 9 & \multicolumn{1}{c|}{4.23} & \textbf{2.7}  & \textbf{35.86} & \multicolumn{1}{c|}{91.1} & 25.8 & \textbf{430.32} & \textbf{98.3} \\ \hline
\end{tabular}
 \label{tb:sotanm}
\end{table*}

To accurately assess the hardware efficiency of TRIP, we implemented our hardware-optimized TRIP algorithm on the SENECA neuromorphic processor \cite{tang2023seneca}. To compare with state-of-the-art neuromorphic solutions on event-based vision, we used the DvsGesture dataset \cite{amir2017low} to benchmark the performance of our solution in terms of accuracy, latency, energy consumption, and hardware's effective area.

\subsubsection{Hardware-optimized TRIP}

The hardware-aware optimizations for TRIP include sparsity-aware training on event-based CNNs, quantization-aware training to get network parameters in low precision, and utilizing DAP for ROI generation. The optimization process comprises three steps. First, we performed sparsity-aware training on pre-trained networks in TRIP to reduce the number of activations in event-based CNNs. Second, we conducted quantization-aware training on the ROI prediction network. The incremental quantization-aware training iteratively quantizes and trains each layer with the straight-through gradient estimator. The training freezes optimally quantized layers and trains the remaining layers. Third, we substituted the truncated Gaussian kernels with DAP and fine-tuned the classification network with incremental quantization-aware training. We quantized all network parameters to 4-bit. The hardware-aware optimizations have minimal influence on accuracy, achieving 98.3\% accuracy on the best model for the DvsGesture dataset, only 0.3\% reduction compared to the best model without hardware-aware optimizations.

\subsubsection{Hardware Implementation and Benchmarking}

\begin{figure}[t]
  \centering
   \includegraphics[width=1\linewidth]{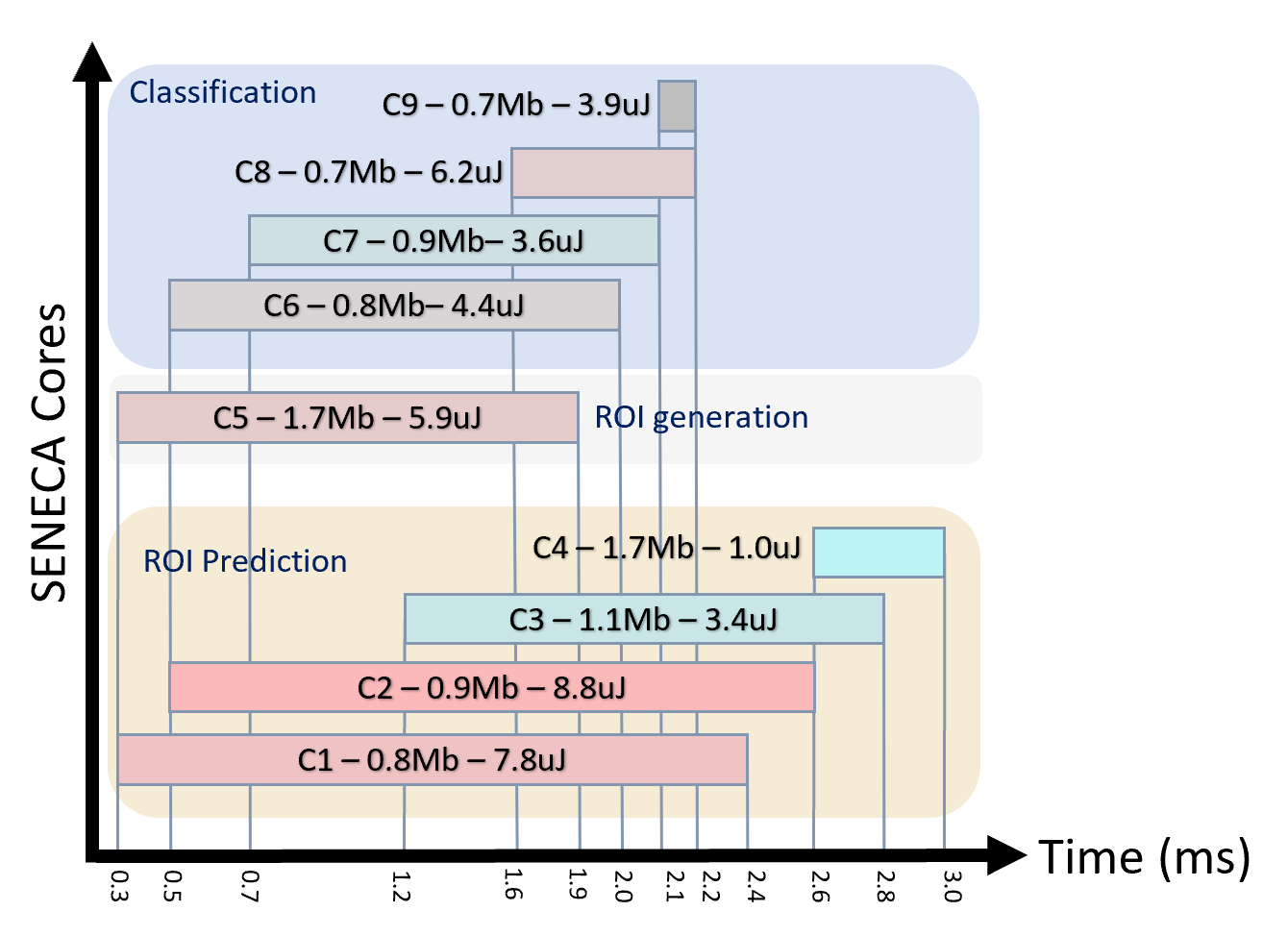}
   \caption{SENECA cores processing pipeline for TRIP}
   \label{fig:util}
\end{figure}

The hardware-optimized TRIP algorithm is implemented on 9 SENECA cores. The ROI prediction network is mapped in 4 cores, including 3 cores for 3 convolutional layers (C1, C2, C3) and one core fusing the ReLU recurrent layer and output layer (C4). The ROI generation with DAP uses a single core (C5). The classification network is mapped in 4 cores, including 2 cores for 2 convolutional layers (C6, C7), one core for the fully-connected layer (C8) and one core for the output layer (C9). Our convolutional layer implementation on SENECA adopts event-driven depth-first convolution \cite{xu18optimizing} and fuses Convolution, BatchNorm, MaxPool, and ReLU on a single core. To improve latency, we parallelized the three components of TRIP on SENECA as shown in Figure \ref{fig:util}. Therefore, the classification is performed on the ROI prediction network's output based on previous timebin inputs. We experimentally observed this approach does not affect the test accuracy. To measure the improvement in SENECA, we created an event-based CNN baseline with the similar number of parameters as two networks in TRIP combined but with higher resolution of inputs ($32\times32$).

\subsubsection{Results}

We benchmarked TRIP on SENECA with state-of-the-art neuromorphic solutions \cite{rueckauer2022nxtf,amir2017low,massa2020efficient} in Table \ref{tb:sotanm}. We presented single and multiple timebins results to enable the comparison with multiple solutions. Single timebin results derive from inferring one timebin of the input event stream, while multiple timebin results reflect maximum accuracy. Our solution outperforms in nearly every metric of the benchmark. Notably, we achieved $46\times$ energy efficiency and $88\times$ area improvements compared to the spiking CNN solution on TrueNorth. One Loihi-based solution shows better latency than ours. However, Loihi's high parallelism results in large areas that are unscalable for high-resolution inputs. Additionally, TRIP decreases the error rate to half and improves energy consumption. Moreover, TRIP improves more than $2\times$ on latency and energy compared to our baseline on SENECA. Since the area is computed by multiplying the used cores with the area/core, TRIP has a higher area cost since uses two additional SENECA cores than the baseline.

To further analyze the behavior of TRIP in the SENECA neuromorphic processor, we visualized the details of each SENECA core in Figure \ref{fig:util}, including memory costs, energy costs, and processing durations. The depth-first CNNs make it possible to parallelize the layers in a pipelined fashion within a single timebin, reducing the inference latency. Moreover, by parallellizing the three TRIP components, the ROI prediction and generation introduce much less latency overhead than sequential processing.

\section{Conclusion and Discussion}

This paper presented TRIP, a hard attention framework for efficient event-based vision processing on the neuromorphic processor. The method achieved state-of-the-art classification accuracy in multiple event-based datasets while exhibiting consistent efficiency improvements in both algorithmic analysis and actual hardware implementation. This demonstrates TRIP's effectiveness in connecting hard attention with event-based vision and neuromorphic computing, offering a viable solution for efficient and low-cost high-resolution visual processing on neuromorphic processors.

A recent trend in sensing technologies involves the seamless integration of sensing and processing hardware to create ultra-efficient sensors with inherent in-sensor and near-sensor processing capabilities \cite{zhou2020near,zhou2023computational}. Employing a similar concept, TRIP can trigger new hardware designs of event-based vision sensors. The ROI prediction network can be accelerated inside the sensory chip for near-sensor processing. The tGK and DAP for ROI generation naturally transforms into hardware circuits next to the sensory array of the event-based camera. The specialized hardware design will increase efficiency and vastly decrease the data bandwidth required by the event stream, relieving the communication burden of downstream neuromorphic processors.

We showed here that TRIP performs efficient classification by actively focusing on regions of the input space. The framework naturally extends to event-based vision tasks with higher resolutions than those we have showcased. Emerging high-resolution event-based cameras, like the 1-megapixel Prophesee camera \cite{perot2020learning}, exhibit promising capabilities for addressing complex applications while requiring substantial processing. Our proposed neuromorphic hard attention solution emerges as a compelling alternative to conventional CNN solutions when aiming for an end-to-end event-based vision system tailored for edge applications.

\bibliographystyle{./IEEEtran}
\bibliography{egbib}

\newpage
\onecolumn
\appendix

\section{Experiment Details}
\subsection{DVS Gesture Dataset}

The following section details the network architecture and training parameters used in the DVS Gesture experiments.

\subsubsection{Network Details}

Table \ref{tb:DVS_ROIpred} shows the sequentially ordered layers from input to output of the ROI prediction network used with the DVS Gesture dataset. Similarly, Table \ref{tb:DVS_classif} shows the layer information of the classification network. Every batch normalization layer is followed by a ReLU activation function.

\begin{table}[h!]
\centering
\caption{Layer overview of ROI prediction net, for TRIP network used in DVS Gesture results.}
\begin{tabular}{l|lllll}
\hline
Layer & Input & Channel & Kernel & Padding  & Number of \\
& dimension & dimensions & size & (no. cells) & neurons \\ \hline
Maxpooling & 128$\times$128 & 2 & 8 & 0   & --  \\
Convolutional & 16$\times$16  & 2, 32 & 3 & 1  & --    \\
Maxpooling & 16$\times$16     & 32 & 2   & 0   & --                           \\ 
Batch Normalization & 8$\times$8 & 32  & -- &  --  & --                         \\ 
Convolutional & 8$\times$8 & 32, 64 & 3 & 1  & --    \\
Maxpooling & 8$\times$8  & 64 & 2   & 0   & --                           \\ 
Batch Normalization & 4$\times$4 & 64  & -- &  --  & --                        \\ 
Convolutional & 4$\times$4 & 64, 128 & 3 & 1  & --    \\
Maxpooling & 4$\times$4  & 128 & 2   & 0   & --                           \\ 
Batch Normalization & 2$\times$2 & 128  & -- &  --  & --                        \\ 
ReLU RNN & 512 & -- & -- & -- & 256\\
Fully Connected & 256 & -- & -- & -- & 3\\ \hline

\end{tabular}
 \label{tb:DVS_ROIpred}
\end{table}

\begin{table}[h!]
\centering
\caption{Layer overview of classification net, for TRIP network used in DVS Gesture results.}
\begin{tabular}{l|lllll}
\hline
Layer & Input & Channel & Kernel & Padding  & Number of \\
& dimension & dimensions & size & (no. cells) & neurons \\ \hline
Convolutional & 12$\times$12  & 2, 32 & 3 & 1  & --    \\
Maxpooling & 12$\times$12     & 32 & 2   & 0   & --                           \\ 
Batch Normalization & 6$\times$6 & 32  & -- &  --  & --                         \\ 
Convolutional & 6$\times$6 & 32, 64 & 3 & 1  & --    \\
Maxpooling & 6$\times$6  & 64 & 2   & 0   & --                           \\ 
Batch Normalization & 3$\times$3 & 64  & -- &  --  & --                        \\ 
Fully Connected & 576 & -- & -- & -- & 256\\
Fully Connected & 256 & -- & -- & -- & 11\\ \hline
\end{tabular}
 \label{tb:DVS_classif}
\end{table}

\subsubsection{Training details}

Table \ref{tb:DVS_params} lists the training hyperparameters used in the DVS Gesture experiments. The accuracy reported in the paper is the mean of the best accuracies obtained from five separate experiments with five random parameter initializations. The training dataset is augmented using the torchvision transforms package. The data augmentation randomly scales samples with a scaling factor between 0.6 and 1.0, applies the random perspective transformation with a distortion parameter of 0.5, and randomly rotates samples between 0 and 25 degrees. SpikingJelly preprocesssing is used to split samples by frame into 32 timebins.

\begin{table}[h!]
\centering
\caption{Training parameters used in DVS Gesture results.}
\begin{tabular}{l|l}
\hline
Parameter & Value  \\ \hline
Learning rate & 0.0001 \\
Training dataset size & 1176 \\
Test dataset size  & 288 \\
Training batch size & 32\\
Testing batch size & 32\\
Number of epochs & 1000 \\ 
Optimizer & Adam \\ \hline
\end{tabular}
 \label{tb:DVS_params}
\end{table}

\subsection{Marshalling Signals Dataset}

On the Marshalling Signals dataset, we employ residual connection to concatenate the RNN output to the input of the first FC layer in the classification network for improved learning stability and accuracy. The detailed parameters of the sequentially ordered layers in the ROI prediciton network and in the classification network are listed in Table \ref{tb:MS_ROIpred} and \ref{tb:MS_classif}, respectively. Every batch normalization layer is followed by a ReLU activation function.

\subsubsection{Network Details}

\begin{table}[h!]
\centering
\caption{Layer overview of ROI prediction net, for TRIP network used in Marshalling Signals results.}
\begin{tabular}{l|lllll}
\hline
Layer & Input & Channel & Kernel & Padding  & Number of \\
& dimension & dimensions & size & (no. cells) & neurons \\ \hline
Maxpooling & 346$\times$224 & 2 & 8 & 0   & --  \\
Convolutional & 43$\times$28  & 2, 32 & 3 & 1  & --    \\
Maxpooling & 43$\times$28     & 32 & 2   & 0   & --                           \\ 
Batch Normalization & 21$\times$14 & 32  & -- &  --  & --                         \\ 
Convolutional & 21$\times$14 & 32, 64 & 3 & 1  & --    \\
Maxpooling & 21$\times$14  & 64 & 2   & 0   & --                           \\ 
Batch Normalization & 10$\times$7 & 64  & -- &  --  & --                        \\ 
Convolutional & 10$\times$7 & 64, 128 & 3 & 1  & --    \\
Maxpooling & 10$\times$7  & 128 & 2   & 0   & --                           \\ 
Batch Normalization & 5$\times$3 & 128  & -- &  --  & --                        \\ 
ReLU RNN & 1920 & -- & -- & -- & 512\\
Fully Connected & 512 & -- & -- & -- & 3\\ \hline

\end{tabular}
 \label{tb:MS_ROIpred}
\end{table}

\begin{table}[h!]
\centering
\caption{Layer overview of classification net, for TRIP network used in Marshalling Signals results.}
\begin{tabular}{l|lllll}
\hline
Layer & Input & Channel & Kernel & Padding  & Number of \\
& dimension & dimensions & size & (no. cells) & neurons \\ \hline
Convolutional & 12$\times$12  & 2, 32 & 3 & 1  & --    \\
Maxpooling & 12$\times$12     & 32 & 2   & 0   & --                           \\ 
Batch Normalization & 6$\times$6 & 32  & -- &  --  & --                         \\ 
Convolutional & 6$\times$6 & 32, 64 & 3 & 1  & --    \\
Maxpooling & 6$\times$6  & 64 & 2   & 0   & --                           \\ 
Batch Normalization & 3$\times$3 & 64  & -- &  --  & --                        \\ 
Fully Connected & 576 & -- & -- & -- & 256\\
Fully Connected & 256 & -- & -- & -- & 11\\ \hline
\end{tabular}
 \label{tb:MS_classif}
\end{table}

\subsubsection{Training details}
The training hyperparameters used in the Marhsalling Signals dataset are listed in Table \ref{tb:MS_params}. The accuracy reported in the paper is the best obtained accuracy during a single experiment. The training dataset is augmented using the torchvision transforms package. The data augmentation randomly scales samples with a scaling factor between 0.6 and 1.0, applies the random perspective transformation with a distortion parameter of 0.5, and randomly rotates samples between 0 and 25 degrees.

\begin{table}[h!]
\centering
\caption{Training parameters used in Marshalling Signals results.}
\begin{tabular}{l|l}
\hline
Parameter & Value  \\ \hline
Learning rate & 0.001 \\
Training dataset size & 11,040 \\
Test dataset size  & 930 \\
Training batch size & 128\\
Testing batch size & 128\\
Number of epochs & 1000 \\ 
Optimizer & Adam \\ \hline
\end{tabular}
 \label{tb:MS_params}
\end{table}

\subsection{Synthetic Dataset Based on N-MNIST}

\subsubsection{Network Details}

The layer-by-layer network details of the ROI prediction network and classification network used for the synthetic dataset based on N-MNIST are listed in Table \ref{tb:NMNIST_ROIpred} and Table \ref{tb:NMNIST_classif} respectively. Every convolutional layer is followed by a ReLU activation function. The maxpooling layers in the classification network use a stride equal to 1. The parameters shown in parenthesis refer to parameters used by the $32\times32$ input size TRIP; the $32\times32$ TRIP uses a kernel size 4 (instead of 8 in the $16\times16$ input size TRIP) in the initial downsampling maxpooling layer, and has an input size of 1600 (instead of 256 in the $16\times16$ input size TRIP) in the ReLU RNN. The remaining network parameters are the same for both $32\times32$ and $16\times16$ input size TRIP. 

\begin{table}[h!]
\centering
\caption{Layer overview of ROI prediction net, for TRIP network used in synthetic N-MNIST-based dataset results.}
\begin{tabular}{l|lllll}
\hline
Layer & Input & Channel & Kernel & Padding  & Number of \\
& dimension & dimensions & size & (no. cells) & neurons \\ \hline
Maxpooling & 128$\times$128 & 2 & 8 (4) & 0   & --  \\
Convolutional & 16$\times$16  & 2, 32 & 5 & 1  & --    \\
Maxpooling & 14$\times$14     & 32 & 2   & 0   & --                           \\ 
Convolutional & 7$\times$7 & 32, 64 & 5 & 1  & --    \\
Maxpooling & 5$\times$5  & 64 & 2   & 0   & --                           \\ 
ReLU RNN & 256 (1600) & -- & -- & -- & 256\\
Fully Connected & 256 & -- & -- & -- & 3\\ \hline

\end{tabular}
 \label{tb:NMNIST_ROIpred}
\end{table}

\begin{table}[h!]
\centering
\caption{Layer overview of classification net, for TRIP network used in synthetic N-MNIST based-dataset results.}
\begin{tabular}{l|lllll}
\hline
Layer & Input & Channel & Kernel & Padding  & Number of \\
& dimension & dimensions & size & (no. cells) & neurons \\ \hline
Convolutional & 12$\times$12  & 2, 32 & 5 & 1  & --    \\
Maxpooling & 10$\times$10     & 32 & 2   & 0   & --                           \\ 
Convolutional & 9$\times$9 & 32, 64 & 5 & 1  & --    \\
Maxpooling & 7$\times$7  & 64 & 2   & 0   & --                           \\ 
Fully Connected & 2304 & -- & -- & -- & 10\\\hline
\end{tabular}
 \label{tb:NMNIST_classif}
\end{table}

The network details of the $16\times16$ input baseline network used in the experiments with the synthetic dataset based on N-MNIST are listed in Table \ref{tb:NMNIST_BL16}. The details of the $32\times32$ baseline are listed in Table \ref{tb:NMNIST_BL32}, and the details of the $64\times64$ baseline are listed in Table \ref{tb:NMNIST_BL64}. Every convolutional layer is followed by a ReLU activation function.

\begin{table}[h!]
\centering
\caption{Layer overview $16\times16$ input resolution baseline network used in synthetic N-MNIST-based dataset results.}
\begin{tabular}{l|lllll}
\hline
Layer & Input & Channel & Kernel & Padding  & Number of \\
& dimension & dimensions & size & (no. cells) & neurons \\ \hline
Maxpooling & 128$\times$128 & 2 & 8 & 0   & --  \\
Convolutional & 16$\times$16  & 2, 16 & 5 & 2  & --    \\
Maxpooling & 16$\times$16     & 16 & 2   & 0   & --                           \\ 
Convolutional & 8$\times$8 & 16, 32 & 5 & 2  & --    \\
Maxpooling & 8$\times$8  & 32 & 2   & 0   & --                           \\ 
Convolutional & 4$\times$4 & 32, 64 & 5 & 2  & --    \\
Maxpooling & 4$\times$4  & 64 & 2   & 0   & --                           \\ 
Convolutional & 2$\times$2 & 64, 128 & 5 & 2  & --    \\
Maxpooling & 2$\times$2  & 128 & 2   & 0   & --                           \\ 
ReLU RNN & 128 & -- & -- & -- & 128\\
Fully Connected & 128 & -- & -- & -- & 64\\ 
Fully Connected & 64 & -- & -- & -- & 10\\ 
\hline

\end{tabular}
 \label{tb:NMNIST_BL16}
\end{table}

\begin{table}[h!]
\centering
\caption{Layer overview $32\times32$ input resolution baseline network used in synthetic N-MNIST-based dataset results.}
\begin{tabular}{l|lllll}
\hline
Layer & Input & Channel & Kernel & Padding  & Number of \\
& dimension & dimensions & size & (no. cells) & neurons \\ \hline
Maxpooling & 128$\times$128 & 2 & 4 & 0   & --  \\
Convolutional & 32$\times$32  & 2, 16 & 5 & 2  & --    \\
Maxpooling & 32$\times$32     & 16 & 2   & 0   & --                           \\ 
Convolutional & 16$\times$16 & 16, 32 & 5 & 2  & --    \\
Maxpooling & 16$\times$16  & 32 & 2   & 0   & --                           \\ 
Convolutional & 8$\times$8 & 32, 64 & 5 & 2  & --    \\
Maxpooling & 8$\times$8  & 64 & 2   & 0   & --                           \\ 
Convolutional & 4$\times$4 & 64, 128 & 5 & 2  & --    \\
Maxpooling & 4$\times$4  & 128 & 2   & 0   & --                           \\ 
ReLU RNN & 512 & -- & -- & -- & 384\\
Fully Connected & 384 & -- & -- & -- & 128\\ 
Fully Connected & 128 & -- & -- & -- & 10\\ 
\hline

\end{tabular}
 \label{tb:NMNIST_BL32}
\end{table}

\begin{table}[h!]
\centering
\caption{Layer overview $64\times64$ input resolution baseline network used in synthetic N-MNIST-based dataset results.}
\begin{tabular}{l|lllll}
\hline
Layer & Input & Channel & Kernel & Padding  & Number of \\
& dimension & dimensions & size & (no. cells) & neurons \\ \hline
Maxpooling & 128$\times$128 & 2 & 2 & 0   & --  \\
Convolutional & 64$\times$64  & 2, 16 & 5 & 1  & --    \\
Maxpooling & 62$\times$62     & 16 & 2   & 0   & --                           \\ 
Convolutional & 31$\times$31 & 16, 32 & 5 & 1  & --    \\
Maxpooling & 29$\times$29  & 32 & 2   & 0   & --                           \\ 
Convolutional & 15$\times$15 & 32, 64 & 5 & 1  & --    \\
Maxpooling & 13$\times$13  & 64 & 2   & 0   & --                           \\ 
Convolutional & 6$\times$6 & 64, 128 & 5 & 1  & --    \\
Maxpooling & 4$\times$4  & 128 & 2   & 0   & --                           \\ 
ReLU RNN & 512 & -- & -- & -- & 384\\
Fully Connected & 384 & -- & -- & -- & 128\\ 
Fully Connected & 128 & -- & -- & -- & 10\\ 
\hline

\end{tabular}
 \label{tb:NMNIST_BL64}
\end{table}

\subsubsection{Training details}
The training hyperparameters used in all of the experiments with the synthetic dataset based on N-MNIST are listed in table \ref{tb:NMNIST_params}. The accuracies reported in the paper are the mean of the best accuracies obtained from five different experiments with five random parameter initializations.  

\begin{table}[h!]
\centering
\caption{Training parameters used in synthetic N-MNIST based-dataset results.}
\begin{tabular}{l|l}
\hline
Parameter & Value  \\ \hline
Learning rate & 0.0006 \\
Training dataset size & 50,000 \\
Validation dataset size & 10,000 \\
Test dataset size  & 10,000 \\
Training batch size & 32\\
Testing batch size & 64\\
Number of epochs & 5 \\ 
Optimizer & Adam \\ \hline

\end{tabular}
 \label{tb:NMNIST_params}
\end{table}

\section{HW Benchmarking details}
\subsection{Baseline Network}
Table \ref{tb:HW_baseline} lists the layer-by-layer network parameters of the baseline network implemented on SENECA. Every batch normalization layer is followed by a ReLU activation function.

\begin{table}[h!]
\centering
\caption{Layer overview of baseline network used in the SENECA benchmarking results.}
\begin{tabular}{l|lllll}
\hline
Layer & Input & Channel & Kernel & Padding  & Number of \\
& dimension & dimensions & size & (no. cells) & neurons \\ \hline
Maxpooling & 128$\times$128 & 2 & 4 & 0   & --  \\
Convolutional & 32$\times$32  & 2, 32 & 3 & 1  & --    \\
Maxpooling & 32$\times$32     & 32 & 2   & 0   & --                           \\ 
Batch Normalization & 16$\times$16 & 32  & -- &  --  & --                         \\ 
Convolutional & 16$\times$16 & 32, 64 & 3 & 1  & --    \\
Maxpooling & 16$\times$16  & 64 & 2   & 0   & --                           \\ 
Batch Normalization & 8$\times$8 & 64  & -- &  --  & --                        \\ 
Convolutional & 8$\times$8 & 64, 128 & 3 & 1  & --    \\
Maxpooling & 8$\times$8  & 128 & 2   & 0   & --                           \\ 
Batch Normalization & 4$\times$4 & 128  & -- &  --  & --  \\
Convolutional & 4$\times$4 & 128, 128 & 3 & 1  & --    \\
Maxpooling & 4$\times$4  & 128 & 2   & 0   & --                           \\ 
Batch Normalization & 2$\times$2 & 128  & -- &  --  & -- \\
Convolutional & 2$\times$2 & 128, 128 & 3 & 1  & --    \\
Maxpooling & 2$\times$2  & 128 & 2   & 0   & --                           \\ 
Batch Normalization & 1$\times$1 & 128  & -- &  --  & -- \\ 
ReLU RNN & 128 & -- & -- & -- & 256\\
Fully Connected & 256 & -- & -- & -- & 11\\ \hline

\end{tabular}
 \label{tb:HW_baseline}
\end{table}

\subsection{Hardware Measurement and Comparison}

All hardware-related measurements were performed in gate-level simulation using industry-standard ASIC simulation and power measurement tools (Cadence Xcelium and Cadence JOULES) for GF-$22$nm FDX technology node (in the typical corner $0.8$V and $25$C, no back-biasing). The power results are accurate within 15\% of signoff power and include the total power consumption of the chip, i.e. both dynamic and static power. The latency results are cycle-accurate with a design frequency of 500 MHz. Same with other compared results, we have not included the I/O power consumption and latency in the reported results. In the reference comparison with other chips, Loihi energy results only includes dynamic power and TrueNorth energy result includes the total power.

\section{Sample Visualizations}

\subsection{DVS Gesture}

\begin{figure}[t]
  \centering
   \includegraphics[width=0.6\linewidth]{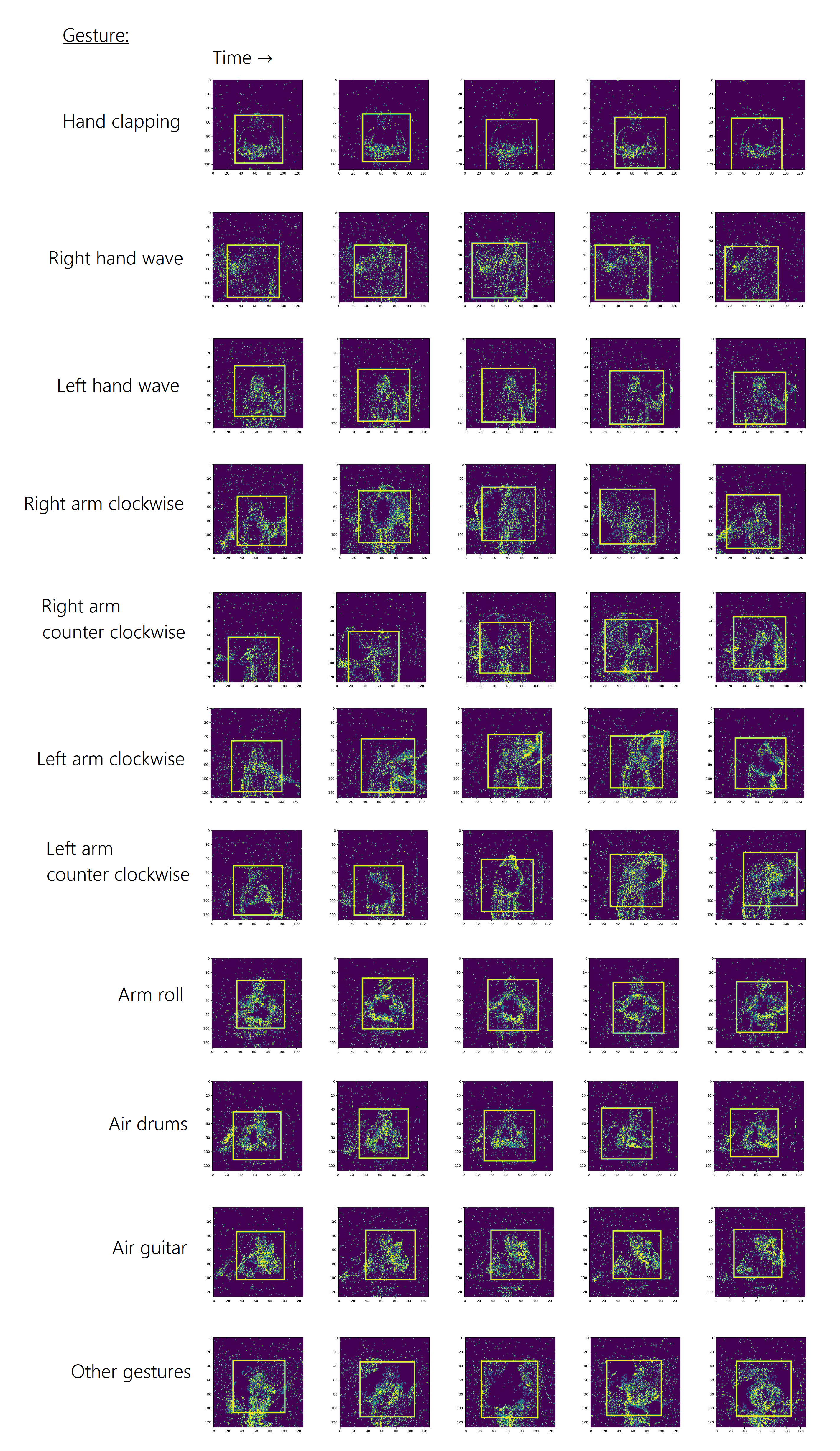}

   \caption{Samples of DVS Gesture dataset with ROI receptive field superimposed as yellow square.}
   \label{fig:DVS_samples}
\end{figure}

Figure \ref{fig:DVS_samples} shows one example per gesture class in the DVS Gesture dataset of a test dataset sample. The first five sequentially ordered timebins from each sample is shown starting from the left, and the ROI receptive field is visualized as a yellow square superimposed on the image.

\subsection{Marshalling Signals}

\begin{figure}[t]
  \centering
   \includegraphics[width=0.6\linewidth]{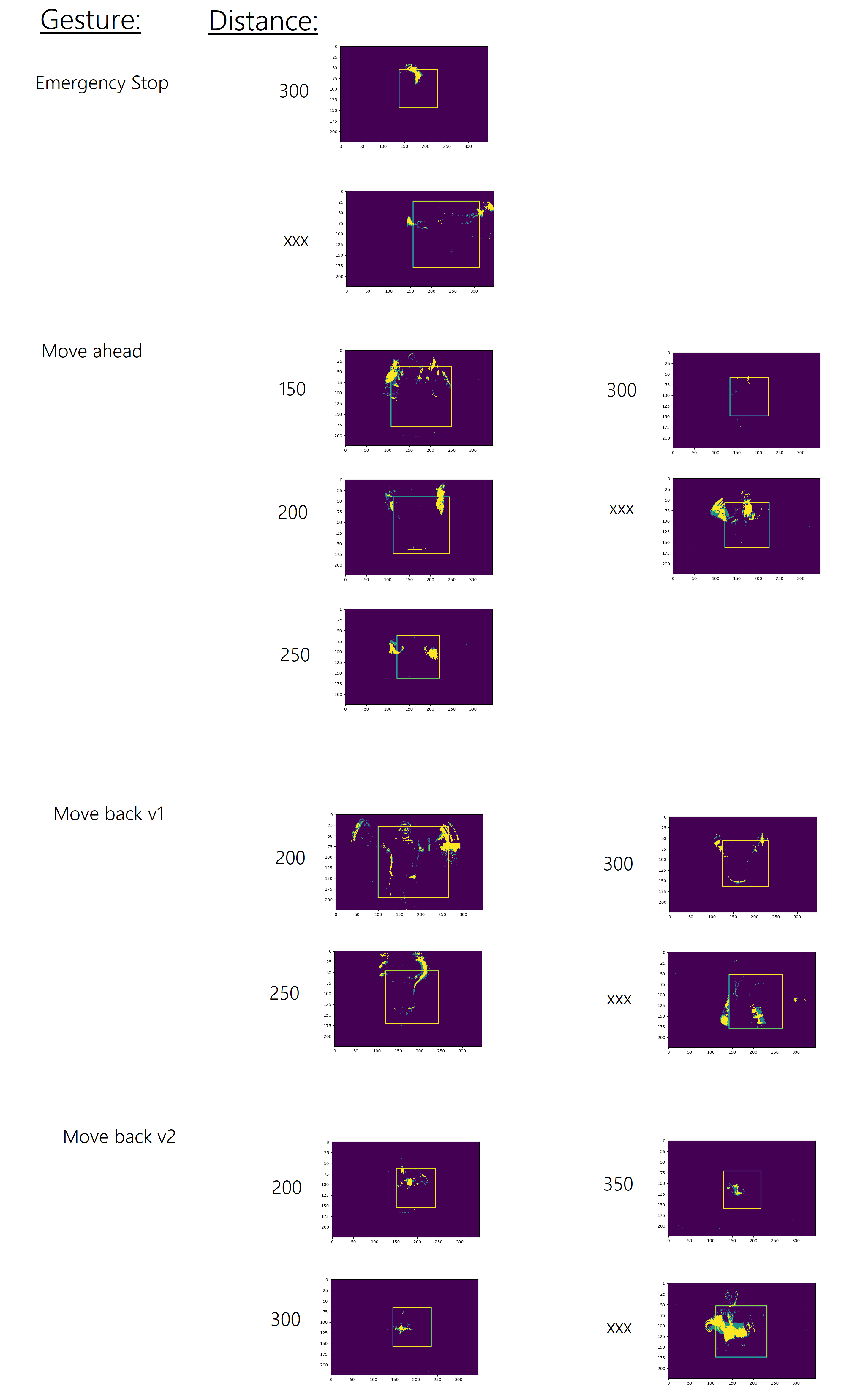}

   \caption{Samples of Marshalling Signals with ROI receptive field superimposed as yellow square.}
   \label{fig:MS_samples_1}
\end{figure}

\begin{figure}[t]
  \centering
   \includegraphics[width=0.4\linewidth]{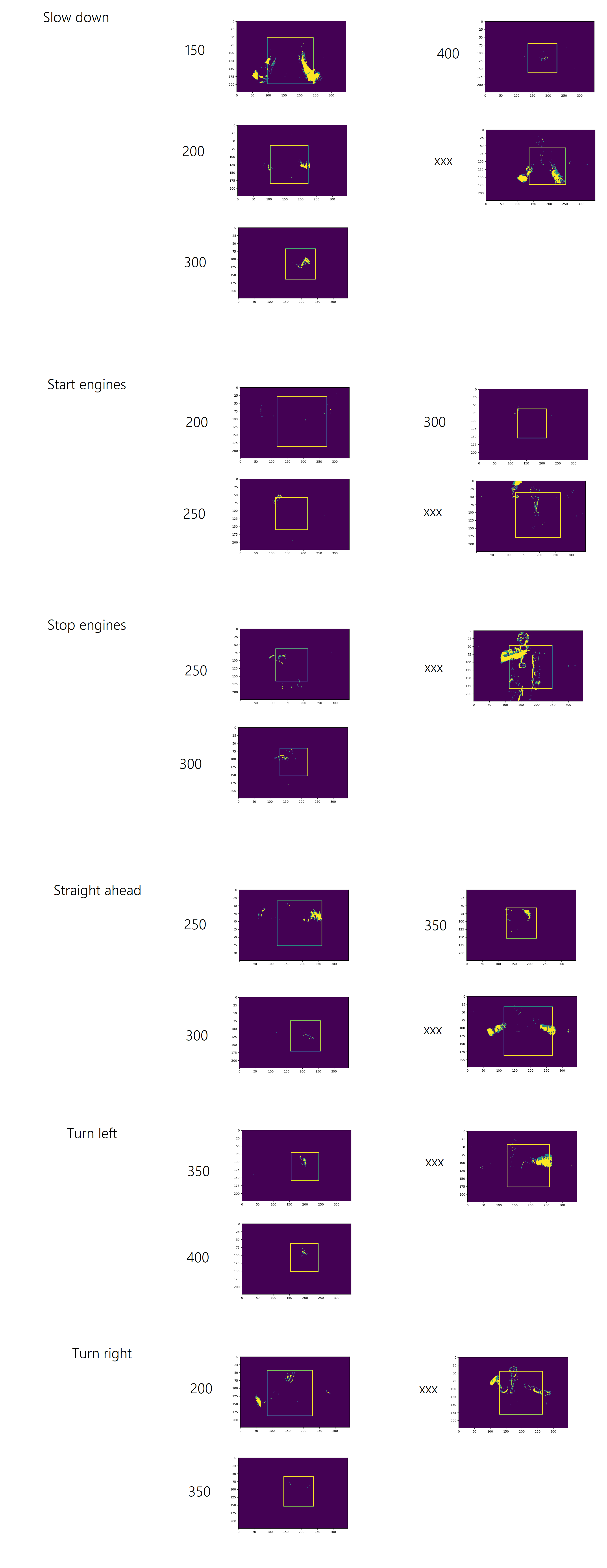}

   \caption{Samples of Marshalling Signals with ROI receptive field superimposed as yellow square.}
   \label{fig:MS_samples_2}
\end{figure}

Example gestures from every class in the Marshalling Signals test dataset are visualized in Figures \ref{fig:MS_samples_1} and \ref{fig:MS_samples_2}. The test dataset does not contain every possible combination of distance and gesture; every unique combination that occurs is shown in the figures. The distance labels indicate the number of centimeters from the camera which the gesture was recorded from. The gestures with distance label "xxx" are samples from real-world output distribution data with unknown distance.

\subsection{Synthetic Dataset Based on N-MNIST}

\begin{figure}[t]
  \centering
   \includegraphics[width=0.6\linewidth]{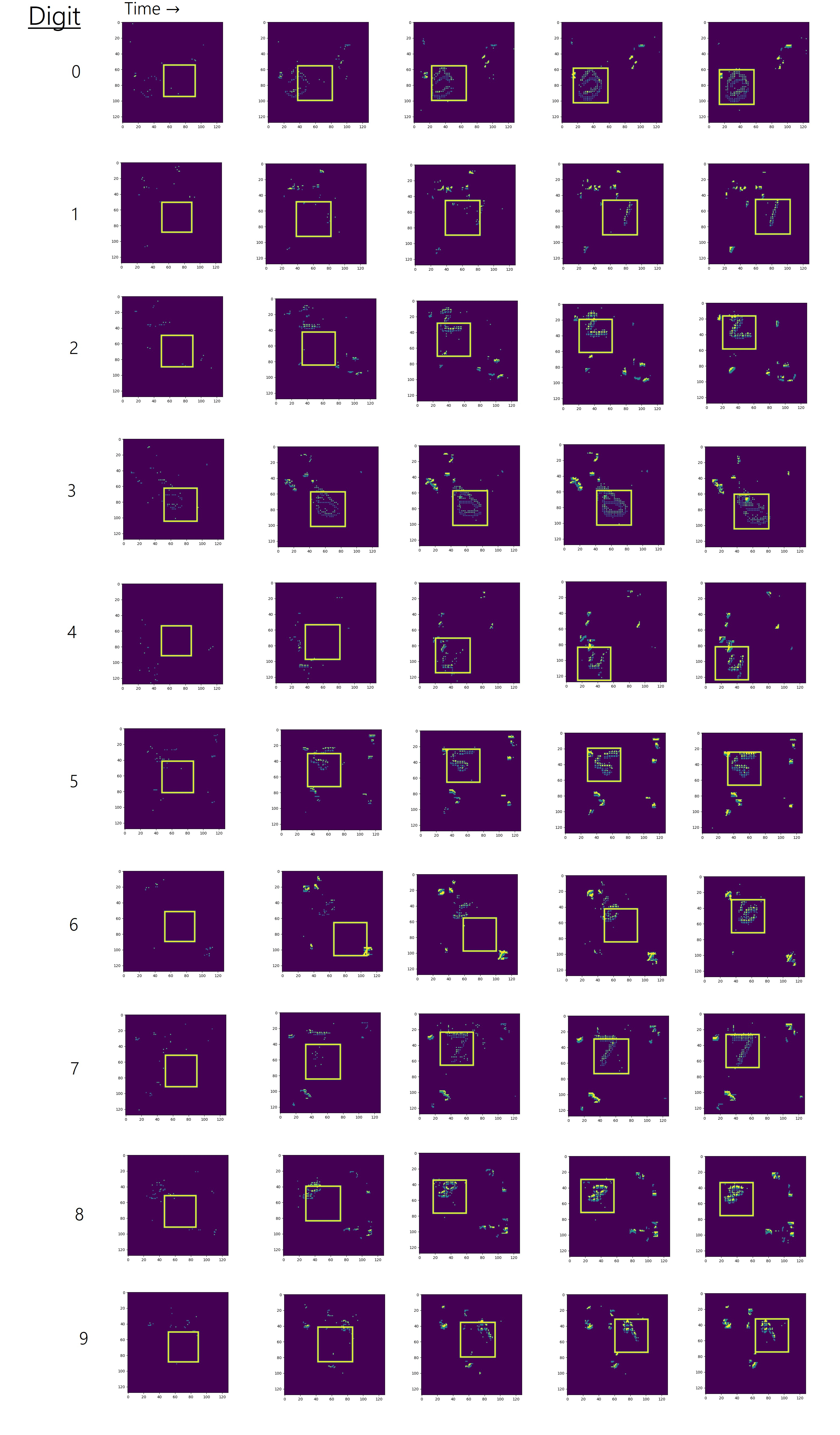}

   \caption{Samples of synthetic dataset based on N-MNIST with ROI receptive field superimposed as yellow square.}
   \label{fig:NMNIST_samples}
\end{figure}

An example testing sample from each digit class of the synthetic dataset based on N-MNIST is visualized in Figure \ref{fig:NMNIST_samples}, together with the ROI receptive field as a yellow superimposed square. The N-MNIST dataset is recorded in such a way that the digit disappears and re-appears between timebins. The first timebins of a sample are empty and the digit cannot be seen until it appears a few timebins later. It can be noted in Figure \ref{fig:NMNIST_samples} that the ROI receptive field initially locates itself somewhere in the center, and as soon as the digit begins to appear it locates the ROI receptive field on the location of the digit. In the case of digit 6, a piece of structured noise appears before the digit 6, and the receptive field begins moving towards the noise. However, once the digit has appeared, the receptive field changes direction and moves towards the digit instead.

\end{document}